%% file: Main.tex
\RequirePackage{xcolor}
\documentclass{article}
\usepackage[final]{corl_2020} % Uncomment for the camera-ready ``final'' version.
\usepackage{graphicx}
\usepackage{todonotes}
\usepackage{booktabs}
\usepackage{amsmath}
\usepackage{hyperref}
\usepackage{amssymb}
\DeclareMathOperator*{\argmax}{\arg\!\max}
\usepackage[ruled,vlined]{algorithm2e}
\usepackage{textcomp}
\usepackage{gensymb}

\title{Fast robust peg-in-hole insertion \\with continuous visual servoing}

\author{
  Rasmus Laurvig Haugaard\\
  SDU Robotics\\
  University of Southern Denmark
  Denmark\\
  \texttt{rlha@mmmi.sdu.dk} \\
  \And
  Jeppe Langaa\\
  SDU Robotics\\
  University of Southern Denmark
  Denmark\\
  \texttt{jela@mmmi.sdu.dk} \\
  \And
    Christoffer Sloth\\
  SDU Robotics\\
  University of Southern Denmark
  Denmark\\
  \texttt{chsl@mmmi.sdu.dk} \\
  \And
  Anders Glent Buch\\
  SDU Robotics\\
  University of Southern Denmark
  Denmark\\
  \texttt{anbu@mmmi.sdu.dk} \\
}

\begin{document}
\maketitle

%===============================================================================

\begin{abstract}
    % The purpose of this document is to provide both the basic paper template and submission guidelines. Abstracts should be a single paragraph, between 4--6 sentences long, ideally. Gross violations will trigger corrections at the camera-ready phase.
    This paper demonstrates a visual servoing method which is robust towards uncertainties related to system calibration and grasping, while significantly reducing the peg-in-hole time compared to classical methods and recent attempts based on deep learning.
    The proposed visual servoing method is based on peg and hole point estimates from a deep neural network in a multi-cam setup, where the model is trained on purely synthetic data.
    Empirical results show that the learnt model generalizes to the real world, allowing for higher success rates and lower cycle times than existing approaches.
\end{abstract}


%===============================================================================
\input{Introduction.tex}
\input{RelatedWork.tex}
\input{Methods.tex}
\input{Results.tex}

\input{Conclusion.tex}
%===============================================================================

% The maximum paper length is 8 pages excluding references and acknowledgements, and 10 pages including references and acknowledgements
\clearpage
% The acknowledgments are automatically included only in the final version of the paper.
% If a paper is accepted, the final camera-ready version will (and probably should) include acknowledgments. All acknowledgments go at the end of the paper, including thanks to reviewers who gave useful comments, to colleagues who contributed to the ideas, and to funding agencies and corporate sponsors that provided financial support.
\acknowledgments{
The authors gratefully acknowledge the economic support from Innovation Fund Denmark through the project MADE FAST.
}

\bibliography{refs.bib}  % .bib
\end{document}

%% file: Introduction.tex
\section{Introduction}
% uncertainties in robotic systems are inherent for desired features of robotic systems
% but high precision tasks forces us to reduce the uncertainty
% we don't want to compromise on the features that introduce the uncertainty, so instead, we handle the uncertainty locally (since it's a learning conference, a human analogy might be fitting here)
% different sensors reduce different uncertainties in different cases
% we focus on peg-in-hole, since it's a common industrial high-precision task
% we note that compliant insertion handles uncertainties well 
%   when peg-hole interaction is achieved (very low uncertainties)
% spiral search and random search handles slighty larger uncertainties well
% 
% 

Uncertainties in robotic systems accumulate from various system-parts, including robot kinematics, tools, object grasping, camera calibrations, etc.
In high-precision tasks, some uncertainties between target objects may have to be reduced.
One approach to reduce the uncertainty is better global calibration, but that is likely to be in conflict with other desired system features, like
    passive grippers,
    passive compliance,
    or 
    modular tool systems which enable more agile systems, but also introduce further uncertainties.
Another approach to reduce uncertainties which remedies the above limitations is to handle the uncertainty live and locally, e.g. with force and/or visual feedback, depending on the task.

While many high-precision tasks exist, we focus on a common industrial task, inserting pegs into holes (peg-in-hole).
Compliant insertion based on force feedback is effective, when the uncertainty is low enough to enable rich peg-hole interaction.
If the uncertainty is slightly larger, classical methods like random search and spiral search can effectively reduce the uncertainty to enable compliant insertion.
When the uncertainty becomes larger, the classical search methods become slow, as they blindly search the uncertainty region.
More importantly, they assume that the surface is flat around the hole within the uncertainty region, which is only a good assumption if the region is small, and are likely to fail otherwise.
In addition, these methods can be harmful for the surface quality of both the peg and the hole-object, and they might result in dropping the peg, if the grasp-force is inadequate, which is not unlikely with passive grippers.

Visual feedback has the potential to effectively reduce larger uncertainties.
Only detecting one of the target objects though, only reduces some of the uncertainties.
E.g. in the peg-in-hole case, 
    mounting in-hand cameras on the peg-robot to detect the hole in the robot's tool-frame will ideally eliminate uncertainties from robot kinematics and the hole position relative to the peg-robot base, but it will not reduce the uncertainties from tooling and grasping but instead introduce new uncertainties from camera calibration.
Classical vision methods tend to either be sensitive to lighting and background clutter or require visual markers.

We propose learnt visual servoing for the initial peg-hole alignment which is robust to system calibration, grasping uncertainties, surface geometry and lighting conditions, while being significantly faster than spiral search, random search and recent methods based on deep learning.
The servoing is based on peg and hole point estimates from a deep neural network in a multi-cam setup, where the model is trained on purely synthetic data.

\begin{figure}
    \centering
    \includegraphics[width=0.9\linewidth]{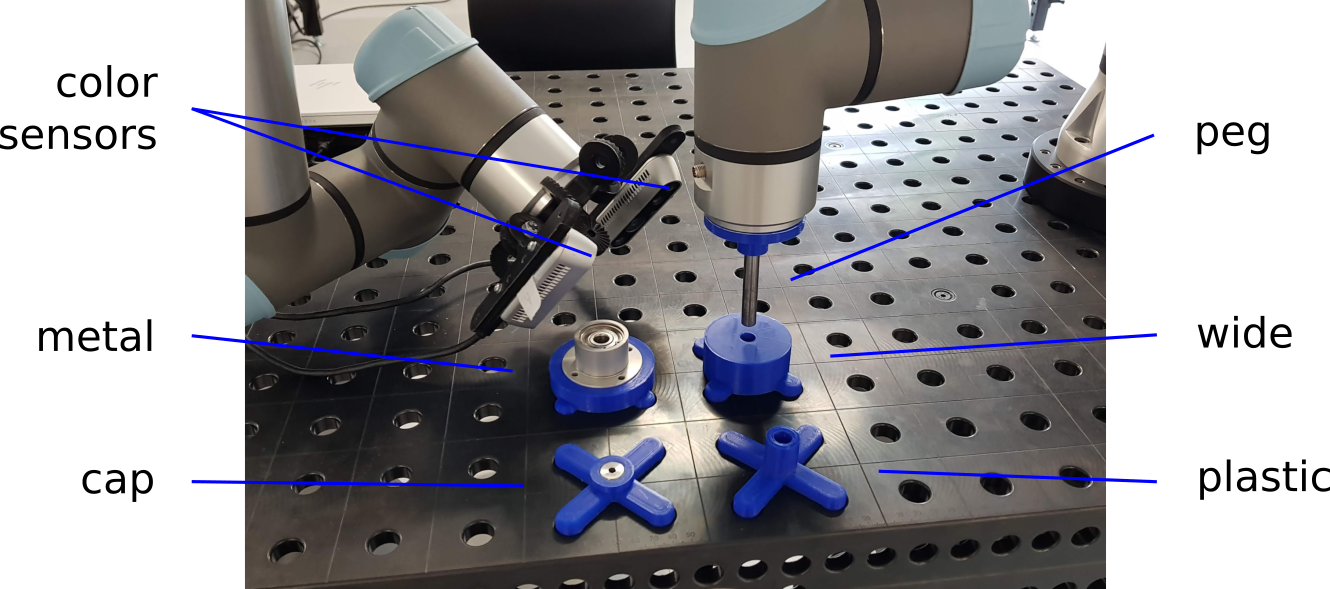}
    \caption{
        Setup consisting of two UR5 robots where 
            two Intel Realsense D435 are mounted on one robot (camera-robot) 
            and a peg is mounted on the other robot (peg-robot).
        Four hole-types, \textit{metal}, \textit{plastic}, \textit{wide} and \textit{cap} are mounted to the table.
        For the \textit{cap} hole, an M4 bolt is used as the peg.
        Note that we only use the color sensors, and thus the size of the camera tool can be greatly reduced.
    }
    \label{fig:system-overview}
\end{figure}

Figure~\ref{fig:system-overview} shows our setup consisting of two cameras mounted on one robot (camera-robot), a peg mounted on another robot (peg-robot), and four holes. 
Our method also allows the cameras to be attached to the peg-robot or be fixed to the table, but having the cameras on a separate robot provides high flexibility, enabling visual servoing for multiple tools on multiple robots.
To test generalization, the setup includes two peg types and four hole types, covering multiple materials, scales and tolerances, as well as different peg and hole surface geometries.

%The implementation will be available online for reproducibility.

%\begin{figure}[!htb]
%	\centering
%	\def\svgwidth{\width=0.9\linewidth}
%    %\def\svgscale{1}
%	\graphicspath{{media/}}
%	\input{media/system-overview.png}
%\caption{
%        Setup consisting of two UR5 robots where 
%            two Intel Realsense D435 are mounted on one robot (camera-robot) 
%            and a peg is mounted on the other robot (peg-robot).
%        Four hole-types, \textit{metal}, \textit{plastic}, \textit{wide} and \textit{cap} are mounted to the table.
%        For the \textit{cap} hole, an M4 bolt is used as the peg.
%        Note that we only use the color sensors, and thus the size of the camera tool can be greatly reduced.
%    }
%    \label{fig:system-overview}
%\end{figure}

%% file: RelatedWork.tex
\section{Related Work}
\label{sec:related-work}

A vast amount of robotic applications rely mainly on predefined and static movements, which do not account for dynamical changes of the environment or task. In contrast, visual servoing incorporates a visual feedback into the control loop, which enables the robot to deal with large deviations. A variety of different visual servoing methods exists including image-based visual servoing \cite{1087115}, pose-based visual servoing \cite{1014775} and hybrid approaches \cite{760345} that use parts of both. Visual servoing opens up for a lot of possibilities by providing feedback without tactility, e.g. tracking and grasping objects with a robot \cite{238279, 7942738}.
Classical visual servoing requires that the robot is accurately calibrated, while more recent methods have tried to address this problem with uncalibrated visual servoing methods \cite{1266656, doi:10.1002/asjc.1756, 9090147}, but these methods needs more time to converge, due to the omitted models of the kinematics and camera calibrations. Instead, we propose to make an initial calibration of the system, but to be robust towards uncertainties in the calibration, similar to \cite{846441}. This ensures that small changes to the setup do not affect the performance of the method, while still exploiting the advantages of a calibrated system.

In deep learning, domain randomization has enabled training models on synthetic data and successfully transferring them to the real world without further training in a variety of tasks, including object detection \cite{annotation-saved}, using a robotic hand to manipulate a cube with reinforcement learning \cite{openai2018learning}, car detection and pose estimation \cite{khirodkar2018domain} and affordance learning \cite{affordance-learning}.
\cite{annotation-saved, khirodkar2018domain, affordance-learning} use synthetic 3D distractor objects to improve their model's generalization. Instead, we overlay natural images on our synthetic images in order to improve generalization.

More closely related to our proposed method is \cite{Huang2013}, which applies visual servoing for peg-in-hole alignment assuming a full pose- estimation from classical methods, where peg and hole are clearly marked. Although the full pose enables the method to handle the rotational alignment of the peg, the requirement of peg and hole markers make such methods undesirable in most use cases. Our method does not require visual markers.

Recently, \cite{quickly-inserting-pegs} proposed image-based visual servoing for peg and hole alignment prior to insertion, which is similar to our approach.
They fix two in-hand cameras to the peg-robot in a top-view configuration and train a deep neural network to classify if the robot should move in one of four directions, or whether they should start the insertion. Their visual servoing is step-wise, not continuous. While they outperform spiral search in their results, forming the problem as classification and using step-wise visual servoing makes the method too slow for practical use with insertion durations in the range of 20-70 seconds. Also, since they only detect the hole, their method does not handle grasping and tool uncertainties.
In contrast, we estimate the positional error of the peg-robot based on both peg and hole point estimates, which handles grasping and tool uncertainties, and reduce the error continuously, enabling significantly faster insertions.

%% file: Methods.tex
\section{Methods}
\label{sec:methods}
% method summary before subsections
This section presents 
    an overview of the alignment methods examined in this work, including
    the proposed visual servoing.
        Our visual servoing alignment is divided into two tasks:
            First, the hole and peg center points are estimated in images using a deep neural network.
            Secondly, based on the point estimates, the peg is aligned to the hole in the plane that is perpendicular to the insertion direction. 
    After alignment, the peg is inserted with compliance using force-feedback.

\subsection{Overview of peg-in-hole alignment}
\label{sec:overview-alignment-methods}
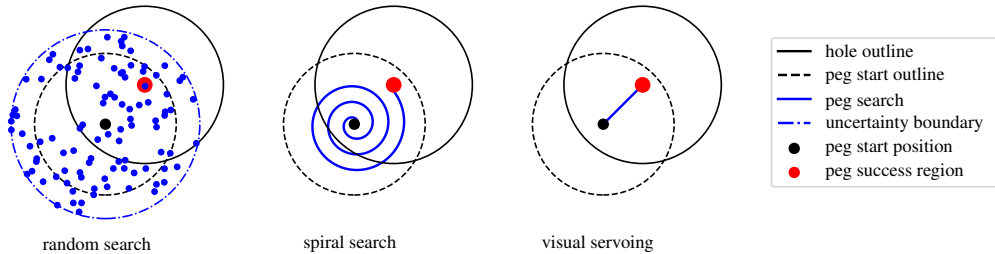
\begin{figure}[!htb]
	\centering
	\def\svgwidth{\columnwidth}
	\graphicspath{{media/}}
{
    \scriptsize
	\input{media/vis_method_overview.tex}
}
\caption{
        Illustration of the three examined methods for peg-in-hole alignment.
        The paths between the points in random search are left out for clarity.
        Note that the peg is in contact with the hole-surface at the search points in random search and throughout spiral search while the peg moves above the surface in visual servoing.
    }
    \label{fig:vis-method-overview}
\end{figure}

We examine three methods for peg-hole positional alignment: random search, spiral search and our proposed visual servoing alignment. The methods are illustrated in Figure~\ref{fig:vis-method-overview}. Note that all three methods only align the position, not the orientation.
In random search, a point within the uncertainty boundary is sampled and an insertion is attempted at that point. This is repeated until the peg reaches below the hole-surface or a time limit is exceeded.
In spiral search, the peg moves outwards in a spiral while pressing against the hole-surface. The peg is partially inserted in the hole when it moves over the hole within a success region based on the peg- and hole tolerances. Spiral search will continue until a force limit is reached, indicating that the peg is in contact with the hole, or until the peg exceeds the uncertainty boundary.
In our proposed visual servoing, both peg and hole center points are continuously estimated in two or more cameras.
An error vector is estimated based on the estimated points, and the error is attempted minimized in a servoing loop.
Visual servoing will continue either until convergence, until a time limit is reached or until the peg exceeds the uncertainty boundary. At convergence, the peg is moved downward along the insertion direction until force-feedback indicates contact. To further increase robustness during alignment, visual servoing can be followed by spiral search or random search with a smaller uncertainty region where appropriate.

After successful positional alignment, there may still be small position uncertainties as well as orientation uncertainties.
To remedy this, the peg is inserted with compliance using force-feedback.

% \begin{figure}
%     \centering
%     \includegraphics[width=\linewidth]{media/vis_method_overview.png}
%     \caption{
%         Illustration of the three examined methods for peg-in-hole alignment.
%         The paths between the points in random search are left out for clarity.
%         Note that the peg is in contact with the hole-surface at the search points in %random search and throughout spiral search while the peg moves above the surface in %visual servoing.
%     }
%     \label{fig:vis-method-overview}
% \end{figure}

\subsection{Point estimation}
We make point estimates based on heatmaps from deep neural networks, similar to \cite{simple-baselines, HRNet} which estimate human pose keypoints, but here adapted to pegs and holes. As shown by \cite{point-regression}, outputting heatmaps allows for better spatial generalization than directly regressing point coordinates from a global, latent representation.
Let $p = (x, y)^T$ be the pixel coordinates of a point in an image, $I$.
Then a desired heatmap, $\Phi$, from a desired point, $p^*$, can be defined using a Gaussian kernel:
\begin{equation}
    \Phi_{p^*} = \exp \left( -\frac{||p - p^*||^2}{2\sigma^2} \right),
\end{equation}
where $\sigma$ is a hyper parameter controlling the size of the active area of the heatmap. 
We set $\sigma = 3 \text{px}$ and have separate heatmaps for the peg and hole points. Figure~\ref{fig:synth-image-example}.f shows an example of target heatmaps.

A deep neural network, $f$, with trainable parameters, $\phi$, estimates the heatmap, $\hat \Phi = f_\phi(I)$, with the same resolution as the input image.
We choose a U-Net \cite{unet} architecture with an ImageNet pre-trained ResNet18 \cite{resnet} backbone for simplicity and to leverage transfer learning while keeping the inference time low. As will be apparent from the results in Section~\ref{sec:result}, this architecture is sufficient to model the problem at hand.
    
    %\rlha{slettet tekst:
    %    \textit{There are more advanced architectures, like \cite{HRNet}, that obtain higher performance on popular point estimation datasets.
    %    They might add a boost to the point estimations but will likely have no effect on the conclusions drawn in this work.}}
    %\ces{Hvorfor?}
    %\rlha{I vores eksperimenter er det ikke vision-modellen, der fejler, så det ville ikke umiddelbart have nogen effekt, at det bliver bedre. Indtil videre har jeg justeret forklaringen, så det ikke er ligeså oplagt at nævne mere avancerede arkitekturer.}

At inference, the point with the largest corresponding value in the estimated heatmap is chosen as the point estimate, $\hat p$:
\begin{equation}
    \hat p = \argmax_p \hat \Phi_p
\end{equation}

\subsubsection{Data generation}
Our proposed model used in our real-world experiments is trained purely on synthetic data with no manual annotations, inspired by \cite{annotation-saved, openai2018learning, khirodkar2018domain, affordance-learning}.
The virtual scene consists of a camera, a cylinder and a surface with a hole in it.
For each image, peg- and camera transforms are sampled, a high dynamic range image (HDRI) environment map is sampled, and a procedural material is sampled. We do not apply randomization to the geometry.

When synthesizing an image, the distance from the camera to the hole is sampled uniformly between 12 cm and 15 cm.
The elevation (the angle between the hole plane and the optical axis) is sampled between 35\degree~and 45\degree. 
The roll angle (the rotation around the optical axis) is sampled between -5\degree and 5\degree.
The peg position is sampled from a disc centered around the hole with a radius of 15 mm and a height relative to the hole between 5 mm and 15 mm.
The peg orientation error is sampled as an axis angle vector, where the axis is sampled from the surface of a unit sphere, and the angle is sampled between 0\degree~and 5\degree.

\begin{figure}
    \centering
    \includegraphics[width=\linewidth]{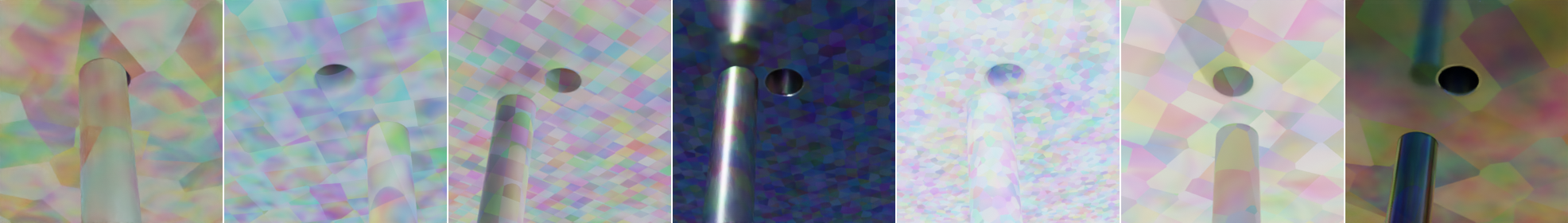}
    \caption{
        Examples of synthetic renders with procedural materials and HDRI environment maps.
    }
    \label{fig:synth-render-examples}
\end{figure}

We use an in-house developed framework~\cite{synth-ml} for synthetic image generation in Blender\footnote{\url{https://www.blender.org/}} with procedural materials and HDRI environment maps from HDRI Haven\footnote{\url{https://hdrihaven.com/}}.
Environment maps provide a large variety of natural lighting and reflections.
We hypothesize that realistic, diverse lighting and reflections enable the model to learn features that generalize well to the real world.
The sampled, procedural material is applied to both the hole-surface and the peg to discourage the model from learning to distinguish between different samples of the procedural material.
We render 1000 images in about 10 minutes on an NVIDIA GeForce RTX 2080.
Examples of the synthetic renders are shown in Figure~\ref{fig:synth-render-examples}.

\begin{figure}
    \centering
    \includegraphics[width=\linewidth]{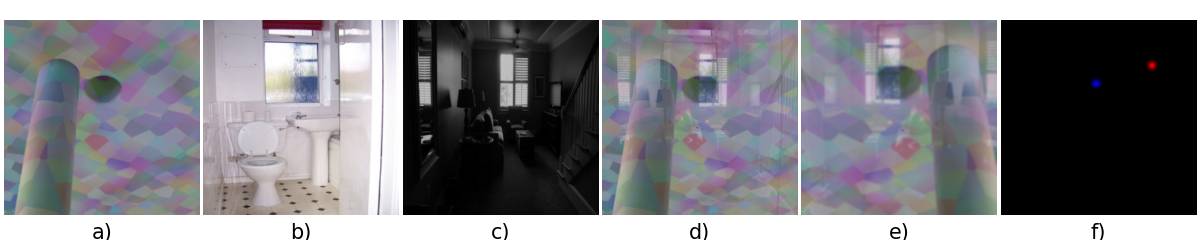}
    \caption{
        Example of a synthetic training image.
        a) Synthetic render.
        b) Overlay.
        c) Mask.
        d) Composite image using a-c.
        e) Composite image with random crop, random horizontal flip, and random blur.
        f) Target heatmaps (hole: blue channel, peg: red channel).
    }
    \label{fig:synth-image-example}
\end{figure}

If the images shown to the model during training only contain the peg and the hole-surface, the model will likely be sensitive to background clutter.
One approach to remedy this is to use natural images as backgrounds as in \cite{synth-blur}, but this encourages the model to discriminate between the natural- and synthetic domain. \cite{synth-blur} alleviates the problem by using somewhat realistic models and freezing a pre-trained backbone.
\cite{annotation-saved, khirodkar2018domain, affordance-learning} overcome the problem by using 3D distractor objects, such that both falses and positives are from the synthetic domain.
Since it is not obvious how to obtain a diverse set of 3D distractor objects, we instead introduce distractors from natural images. 
We sample a natural image as an overlay image and another natural image, converted to grayscale, as the alpha channel of the overlay image.
We sample the natural images from the MS COCO dataset \cite{mscoco}.
The augmentation is illustrated in Figure~\ref{fig:synth-image-example}(a-d).
Since the overlay from the natural domain is imposed on the whole image, the model will not be able to simply distinguish between the synthetic- and natural domain.
Also, natural overlays on peg and hole will effectively serve as extra randomization on peg and hole appearance, which can be hard to simulate.
Since compositing is computationally efficient, the augmentation is done live during training with different overlays at each epoch, effectively reducing the amount of required renders.

\begin{figure}
    \centering
    \includegraphics[width=\linewidth]{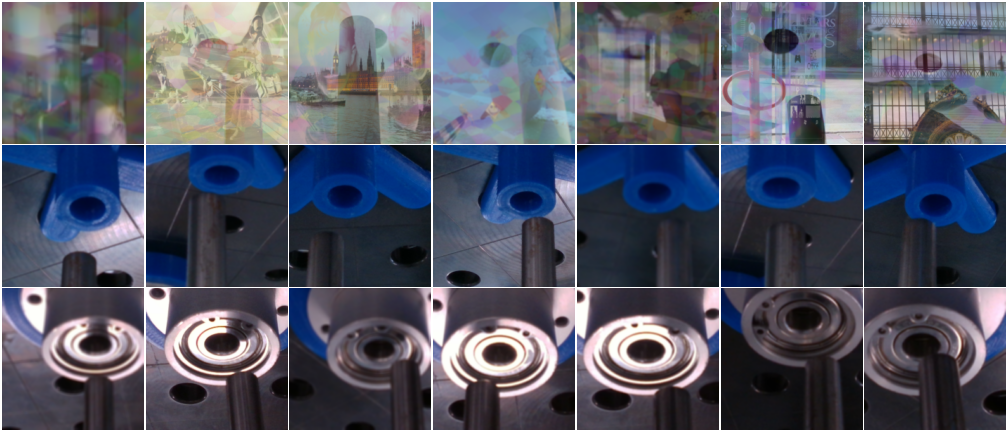}
    \caption{Examples from the three training datasets, from top- to bottom row: synth, plastic, metal.}
    \label{fig:train-image-examples}
\end{figure}

To evaluate the model trained on synthetic images, we obtain natural datasets for the \textit{metal} and \textit{plastic} holes.
To obtain the datasets,
    a rigid camera mount is attached to the camera-robot,
    the whole setup is calibrated,
    and
    robot movements between consecutive sampled peg and camera transforms are ensured to be safe.
The peg and camera transforms are sampled like for the synthetic images. 
We do not attempt to add any variations to the natural environment, e.g. turning on and off lights or capturing images from multiple rooms, since it is impractical.

\subsubsection{Training details}
All images are resized to 224x224 before they are handed to the model.
The loss is defined as the mean squared error between the desired and output heatmaps.
Each of our datasets consists of 1000 images, where 100 of them are reserved for validation.
We also trained on 10x more images on the synthetic domain while keeping the total amount of parameter updates, but with little to no effect. We conclude that the data augmentation provides enough variation within the domain.
We train our models with the one-cycle \cite{onecycle} policy for 30 epochs with a maximum learning rate of 0.001 and weight decay set to 0.0001.
During training, both synthetic and natural images are augmented with random crop, random horizontal flip and a 50\% chance of box blur with a kernel size of either 3 or 5 px.
    The augmentation is illustrated for a synthetic image in Figure~\ref{fig:synth-image-example}.
One training takes about 8 minutes on an NVIDIA GeForce RTX 2080,
and we obtain single-image inference at 150 Hz.

\subsection{Visual servoing}
Our method requires regions of interest (ROIs) in the images, providing crops as seen in Figure~\ref{fig:train-image-examples}, as well as approximated depths (distance from cameras to peg and hole).
    ROIs and depths can either be calculated based on an approximate hole frame or by manually marking the ROIs in the full images once for a given setup.
Our visual servoing alignment method is outlined in Algorithm~\ref{alg:visual-servo}.

\begin{algorithm}
\DontPrintSemicolon
\SetAlgoLined
    \KwIn{
        Insertion direction, $l$.
        Estimated depths, $z_i$, in $n$ cameras.
        Target convergence threshold, $\phi_t$.
        Filter parameters $\alpha_\tau$, $\alpha_\gamma$ and $\alpha_\phi$.
    }
    $\tau \gets$ initial peg-robot position\tcp*{filtered target position}
    $\phi \gets 10\phi_t$\tcp*{filtered error magnitude}
    \While{$\phi > \phi_t$}{
        $q\gets$get position of peg-robot\;
        \For{$i = 1, ..., n$}{
            $I_i \gets$ get image from camera $i$\;
            $p_i^*, h_i^* \gets$ estimate peg and hole points in $I_i$\;
            $p_i, h_i \gets$ estimate 3D points based on $p_i^*, h_i^*$, $z_i$ and the system state- and calibration\;
            $c_i \gets$ get position of camera $i$\;
            $v_i \gets (p_i + h_i) / 2 - c_i$\tcp*{view direction}
            $u_i \gets \dfrac{v_i \times l}{|| v_i \times l ||}$\tcp*{direction of error from this view}
            $b_i  \gets u_i \cdot (h_i - p_i)$\tcp*{magnitude of error from this view}
        }
        $(A, b) \gets \left( 
            \begin{bmatrix}
                u_{1x} & u_{1y} & u_{1z} \\
                u_{2x} & u_{2y} & u_{2z} \\
                & \vdots
            \end{bmatrix}
            ,
            \begin{bmatrix}
                b_1 \\ b_2 \\ \vdots
            \end{bmatrix}
            \right)
        $\;
        $\hat{e} \gets$ solve $(A, b)$ by least squares\tcp*{estimated error vector}
        $t \gets q + \hat{e}$\tcp*{unfiltered target position}
        $\tau \gets \alpha_\tau \tau + (1 - \alpha_\tau) t$\tcp*{update filtered target position}
        set peg-robot reference to $\tau$\tcp*{move robot towards $\tau$}
        $\gamma \gets \alpha_\gamma \gamma + (1 - \alpha_\gamma) \hat{e}$ if $\gamma$ is defined, else $\hat{e}$\tcp*{update filtered error vector}
        $\phi \gets \alpha_\phi \phi + (1 - \alpha_\phi) ||\gamma||$\tcp*{update filtered error magnitude}
    }
    \caption{Visual servo positional alignment}
\label{alg:visual-servo}
\end{algorithm}

While performing the alignment with visual servoing, the robot moves in a plane perpendicular to a provided insertion direction, $l$.
    Note that the insertion direction is the known direction in which the peg-robot will move when the positional alignment along the plane is done. It does thus not need to be exactly parallel to the axis of the peg or the hole.
Because of depth-ambiguity, points from the images of the $i$'th camera provide information only along the direction, $u_i$, in the movement plane that is perpendicular to the view direction, $v_i$.
The point estimates from the image are projected to 3D based on the system state, system calibration and constant depth estimates, $z_i$, used for both the peg- and hole point, $p_i, h_i$. 
%We then define the view direction as $v_i = (p_i + h_i)/2 - c_i$, where $c_i$ is the camera position.
%The size of the error, $||e_i||$, along $\hat e_i$ is then:
%\begin{equation}
%    ||e_i|| = \hat e_i \cdot (h_i - p_i)
%\end{equation}
%Note that $\hat e_i$ is independent of $z_i$ and $||e_i||$ is directly proportional to $z_i$.
Peg and hole estimates from multiple cameras then allow us to estimate the peg-robot error vector $\hat e$.
%solver:
%\begin{equation}
%    \begin{bmatrix}
%    \hat e_{1x} & \hat e_{1y} & \hat e_{1z} \\
%    \hat e_{2x} & \hat e_{2y} & \hat e_{2z} \\
%    & \vdots
%    \end{bmatrix}
%    \begin{bmatrix}
%    e_x \\ e_y \\ e_z
%    \end{bmatrix}
%    = 
%    \begin{bmatrix}
%    ||e_1|| \\ ||e_2|| \\ \vdots
%    \end{bmatrix}
%\end{equation}
% image Jacobian

% smoothing and success condition
A target position, $t$, for the peg-robot is found by adding the estimated error vector, $\hat e$, to the current peg-robot position, similar to visual servoing algorithms that use the estimated image Jacobian for controlling the robot. The target position and the error magnitude are filtered as shown in Algorithm~\ref{alg:visual-servo} with coefficients $\alpha_\tau=\alpha_\gamma=\alpha_\phi=0.9$. Finally, we use the built-in servoing api from the UR5 robot to move towards the target position, and terminate the visual servoing when the filtered error magnitude is smaller than the threshold $\phi_t$, which is set to be 1/20 of the peg-hole diameter.
%To avoid instability caused by the delay from image acquisition, model inference and networking, we calculate $t$ with respect to the system state when the images were taken.

%Lastly, the convergence criterion is described.
%Let $\epsilon_t$ be a predefined convergence threshold.
%Let $\hat e$ be the exponentially smoothed $e$ with $\alpha=0.9$,
%$\epsilon = ||\hat e||$, and $\hat \epsilon$ be the exponentially smoothed $\epsilon$ %with $\alpha=0.9$. \agb{(Brug et andet symbol end $\epsilon$)}
%Visual servoing is successful when $\hat \epsilon < \epsilon_t$.
%We set $\epsilon_t$ to be 1/20 of the peg-hole diameter, and we initialize $\epsilon_0 = 10\epsilon_t$ to avoid \agb{spurious early} termination.
%Using the smoothed $\hat \epsilon$ instead of $\epsilon$ prevents termination when $\hat e$ passes by close to the zero-vector, e.g. in case of overshooting.

%\begin{proposition}
%Test
%\end{proposition}

%% file: media/vis_method_overview.tex
%% Creator: Inkscape inkscape 0.92.2, www.inkscape.org
%% Accompanies image file 'vis_method_overview.pdf' (pdf, eps, ps)
\begingroup%
  \makeatletter%
  \providecommand\color[2][]{%
    \errmessage{(Inkscape) Color is used for the text in Inkscape, but the package 'color.sty' is not loaded}%
    \renewcommand\color[2][]{}%
  }%
  \providecommand\transparent[1]{%
    \errmessage{(Inkscape) Transparency is used (non-zero) for the text in Inkscape, but the package 'transparent.sty' is not loaded}%
    \renewcommand\transparent[1]{}%
  }%
  \providecommand\rotatebox[2]{#2}%
  \ifx\svgwidth\undefined%
    \setlength{\unitlength}{648bp}%
    \ifx\svgscale\undefined%
      \relax%
    \else%
      \setlength{\unitlength}{\unitlength * \real{\svgscale}}%
    \fi%
  \else%
    \setlength{\unitlength}{\svgwidth}%
  \fi%
  \global\let\svgwidth\undefined%
  \global\let\svgscale\undefined%
  \makeatother%
  \begin{picture}(1,0.27777778)%
    \put(0,0){\includegraphics[width=\unitlength,page=1]{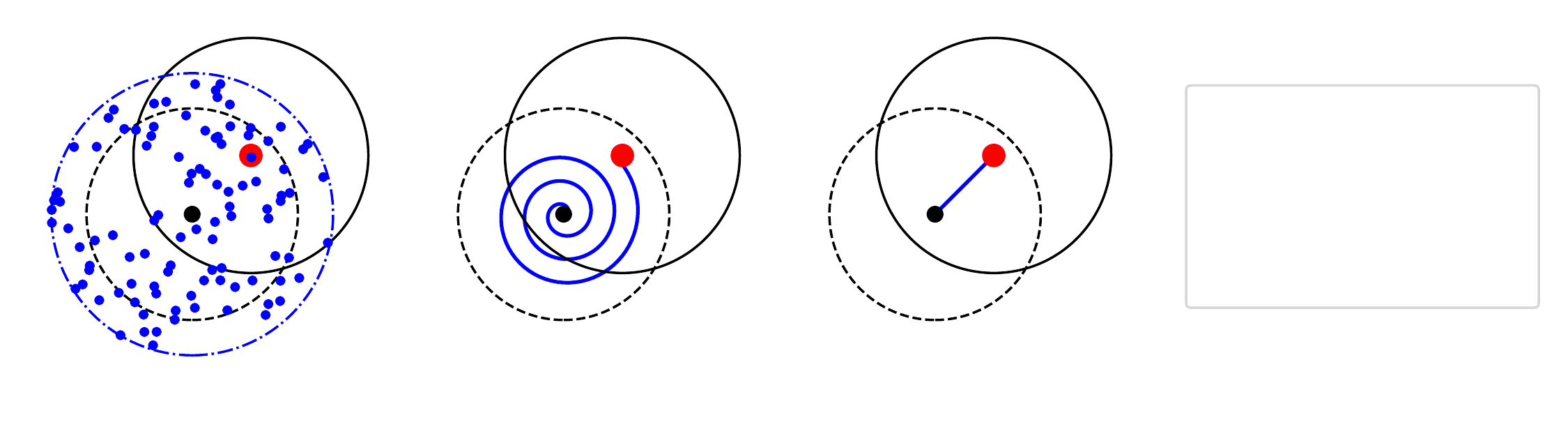}}%
    \put(0.0629845,0.02131782){\color[rgb]{0,0,0}\makebox(0,0)[lb]{\smash{random search}}}%
    \put(0,0){\includegraphics[width=\unitlength,page=2]{vis_method_overview.pdf}}%
    \put(0.5377907,0.0236807){\color[rgb]{0,0,0}\makebox(0,0)[lb]{\smash{visual servoing}}}%
    \put(0.31104653,0.02361439){\color[rgb]{0,0,0}\makebox(0,0)[lb]{\smash{spiral search}}}%
    \put(0.80799856,0.20540943){\color[rgb]{0,0,0}\makebox(0,0)[lb]{\smash{hole outline}}}%
    \put(0.80799819,0.183271){\color[rgb]{0,0,0}\makebox(0,0)[lb]{\smash{peg start outline}}}%
    \put(0.80799819,0.15876954){\color[rgb]{0,0,0}\makebox(0,0)[lb]{\smash{peg search}}}%
    \put(0.80804641,0.13749614){\color[rgb]{0,0,0}\makebox(0,0)[lb]{\smash{uncertainty boundary}}}%
    \put(0.80799819,0.11493995){\color[rgb]{0,0,0}\makebox(0,0)[lb]{\smash{peg start position}}}%
    \put(0.80799819,0.09206911){\color[rgb]{0,0,0}\makebox(0,0)[lb]{\smash{peg success region}}}%
  \end{picture}%
\endgroup%

%% file: Results.tex
\section{Experiments}
\label{sec:result}
In this section, we empirically validate different aspects of our visual servoing system. We start with an accuracy test of the visual part, the point estimation, and then we go on to test the full system, including both alignment and insertion.

\subsection{Point estimation accuracy}
\begin{figure}[ht]
    \centering
    \includegraphics[width=\linewidth]{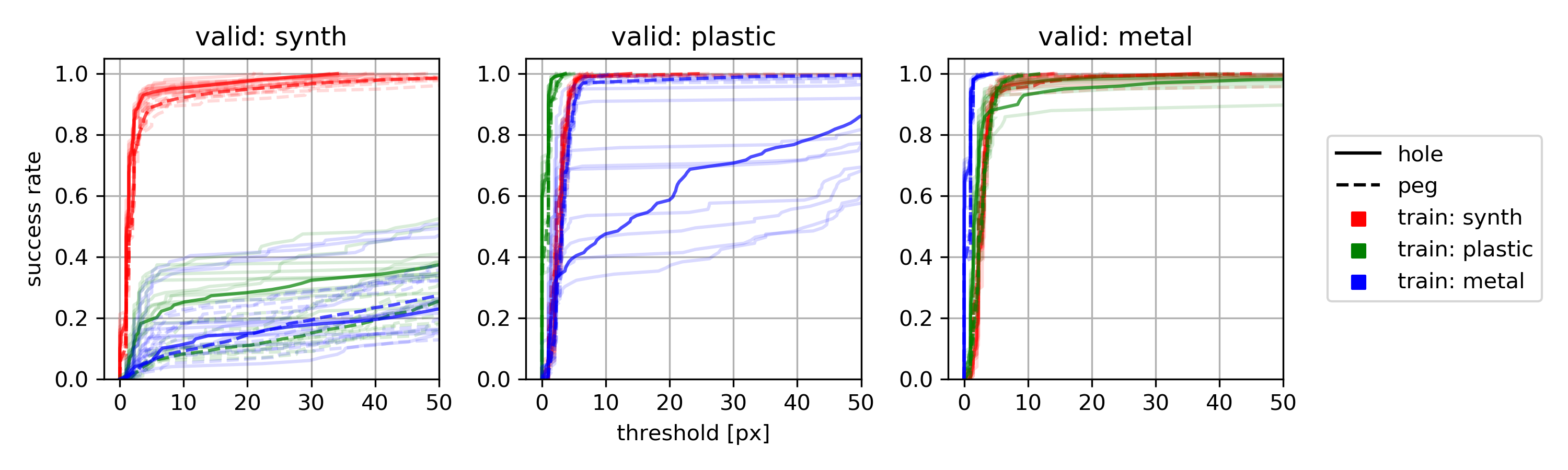}
    \caption{
        Point detection success rates at different pixel thresholds.
        10 models are trained for each of the three training datasets, 
        and each model is evaluated on all of the validation datasets.
        Both mean performance per training dataset and the individual model performances are plotted.
        The individual performances are plotted in a lighter color for clarity.
        Note that the peg- and hole diameter is approximately 50 px in the images.
    }
    \label{fig:synth-vs-real}
\end{figure}

Three models are trained on images of \textit{plastic}, images of \textit{metal} (see Figure~\ref{fig:system-overview}), and the synthetic images, respectively (900 images per dataset).
Each model is evaluated on all validation datasets (100 images per dataset).
The cross-performance is illustrated in Figure~\ref{fig:synth-vs-real}, showing detection success rates versus increasing distance tolerances.
The models trained on the natural domains fail on the synthetic domain, and more importantly, the model trained on the \textit{metal} domain fails to detect the \textit{plastic} hole consistently.
In contrast, the models trained on the synthetic domain perform consistently well in all three domains, indicating that the synthetic images force the models to learn more general features.

\subsection{Peg-in-hole performance}
Using models trained on all three domains, we compare our visual servoing to random search and spiral search.
The results are shown in Table~\ref{tab:performance}
See supplementary material for a video of the methods.

The initial peg positions are sampled from a disc centered around the hole with a diameter of three times the peg diameter and a height relative to the hole between 5 mm and 15 mm (3 mm and 5 mm for \textit{cap}). 
The peg-hole orientation error is sampled up to 2\degree.
To show robustness towards calibration, camera extrinsic errors are sampled with up to 2\degree\ in orientation and 10~mm in position. Note that these are large calibration errors.

The peg used for \textit{bearing}, \textit{plastic} and \textit{wide} is the 10~mm shaft with an h7 tolerance shown in Figure~\ref{fig:system-overview}.
The \textit{bearing} hole is 10~mm with an H7 tolerance while the \textit{plastic} and \textit{wide} holes are 3D-printed with a measured diameter of 10.6~mm and 10.4~mm, respectively.
The \textit{cap} hole is 4.4~mm and the peg is an M4 bolt with a measured diameter of 3.9~mm.

Random search relies on detecting the peg reaching below the hole-surface. For this reason we do not apply randomization in peg height. It also means that random search requires a rather large tolerance compared to the uncertainty, which is why random search is not considered for \textit{metal} with H7/h7 tolerance.
Visual servoing is followed by spiral search as mentioned in Section~\ref{sec:overview-alignment-methods} for \textit{metal}, \textit{plastic} and \textit{wide}, but is avoided for \textit{cap} to avoid damaging the hole surface with the M4 bolt. 
For the same reason, we do not attempt force-based insertion for \textit{cap}.

\begin{figure}
    \centering
    \includegraphics[width=0.19\linewidth]{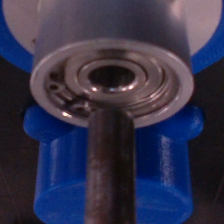}
    \includegraphics[width=0.19\linewidth]{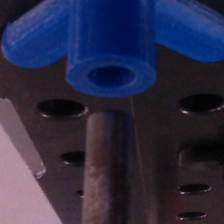}
    \includegraphics[width=0.19\linewidth]{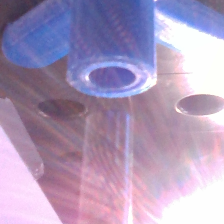}
    \includegraphics[width=0.19\linewidth]{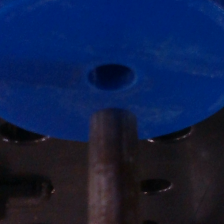}
    \includegraphics[width=0.19\linewidth]{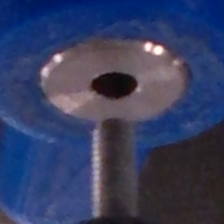}
    \caption{
        Example images from the cameras during visual servoing of the peg-in-hole cases.
        From left to right: \textit{metal}, \textit{plastic}, \textit{plastic*}, \textit{wide} and \textit{cap}.
    }
    \label{fig:heavy_light}
\end{figure}

In the \textit{plastic*} experiment, a light source was placed pointing into the cameras, causing flares and hard reflections on the hole-surface (see Figure~\ref{fig:heavy_light}). All attempts were successful.
The peg for the \textit{cap} hole is threaded, and is thus far from any of the three examined domains.
The models trained on the natural domains completely fail to generalize, while the model trained on the synthetic domain succeeds in all the attempts.

The attempts with our proposed visual servoing are consistently both faster and more robust than random search and spiral search.
Furthermore, the model trained on the synthetic domain generalize to all the examined cases, in contrast to the models trained on natural domains.
\cite{quickly-inserting-pegs} also performs peg-in-hole visual servoing based on deep learning on synthetic images. They use a 10~mm peg and a 10.4~mm hole, similar to \textit{wide}, and a similar but only positional error in the start positions. They achieve insertion times between 20~s and 70~s, while our average insertion time is 2.3~s.

\begin{table}
    \centering
    \footnotesize
    \begin{tabular}{l|ll|lll|l}
        & \multicolumn{2}{c|}{classical search methods} & \multicolumn{3}{c|}{visual servoing, trained on:} \\
                            & random     & spiral     & plastic       & metal         & \textbf{synth (proposed)} & optimal \\ \toprule
        \textit{metal}      & -                 & 10\% (7.1 s)      & \textbf{99\%} (4.1 s)  & \textbf{100\%} (4.0 s) & \textbf{100\%} (3.6 s) & 3.2 s \\
        \textit{plastic}    & 0\% (-)           & 27\% (10 s)     & \textbf{100\%} (2.4s)  & \textbf{100\%} (2.7 s) & \textbf{100\%} (2.2 s) & 1.0 s \\
        \textit{plastic*}   & -                 & -                 & 93\% (2.6 s)  & 11\% (3.7 s)  & \textbf{100\%} (2.3 s) & - \\
        \textit{wide}       & 4\% (15 s)      & \textbf{100\%} (36 s)  & \textbf{99\%} (2.4 s)  & \textbf{100\%} (2.2 s) & \textbf{100\%} (2.3 s) & 1.0 s \\
        \textit{cap}        & 34\% (16 s)     & -                 & 1\% (1.7 s)   & 0\% (-)       & \textbf{100\%} (1.6 s) & 0.4 s \\ \bottomrule
    \end{tabular}
    \medskip
    \caption{
        Success rates and mean time for successful attempts.
        Timings indicate the total time from the initial position to reaching an insertion depth of 10~mm (5~mm for \textit{cap}).
        Each entry is based on 100 runs.
        Highest success rates are marked in bold. 
        If an entry is not significantly lower than the highest success rate with a p-value of 5\%, it is also marked in bold.
        Optimal: peg is aligned to the hole based on calibration (no search).
    }
    \label{tab:performance}
\end{table}

%% file: Conclusion.tex
\section{Conclusion}
\label{sec:conclusion}
We studied the use of deep learning based point estimation and continuous visual servoing for peg-in-hole positional alignment.
We demonstrated that our method can significantly increase speed and robustness compared to classical methods like random search and spiral search.
We also showed that our method is significantly faster than a previous attempt based on deep learning.
Since visual servoing does not require peg-hole contact, our method is robust to hole surface geometry and gentle to the peg and hole surfaces.
We trained point estimation models on both natural and synthetic domains.
We demonstrated that synthetic domain randomization and using distractors from natural images introduces enough variation to enable the model to generalize to all the examined peg-in-hole cases.
While compliant force insertion is likely not enough to account for larger angle errors in tight tolerances,
    integrating axis alignment in visual servoing could remedy this problem and is an area for future work.
Finally, our method is focused on peg-in-hole tasks but we hypothesize that deep learning based visual servoing can be applied successfully in many assembly sub-tasks.